\documentclass{bmvc2k}

\usepackage{amsmath}
\usepackage{booktabs}
\usepackage{array}
\usepackage{placeins}
\usepackage{tikz}
\usetikzlibrary{arrows.meta,positioning}
\usepackage{pgfplots}
\pgfplotsset{compat=1.18}

\newcommand{\ovetc}{OVET-Consensus}

\title{\ovetc{}: Rank-Preserving Counterfactual Event Verification\\
for Video Moment Grounding}

\addauthor{Alibay Osmanli}{a.osmanli@qmul.ac.uk}{1}
\addauthor{Shaogang Gong}{s.gong@qmul.ac.uk}{1}
\addinstitution{Queen Mary University of London\\London, UK}

\runninghead{Osmanli, Gong}{\ovetc{}: Rank-Preserving Event Verification}
\hypersetup{pdftitle={OVET-Consensus: Rank-Preserving Counterfactual Event Verification for Video Moment Grounding},pdfauthor={Alibay Osmanli, Shaogang Gong}}
\makeatletter
\def\Hy@StartlinkName#1#2{\pdfstartlink attr{#1}goto name{#2}\relax}
\def\hyper@natlinkstart#1{\Hy@backout{#1}\hyper@linkstart{cite}{cite.#1}\def\hyper@nat@current{#1}}
\def\hyper@natlinkend{\hyper@linkend}
\makeatother

\begin{document}

\maketitle

\begin{abstract}
Modern video moment retrieval systems often propose windows that overlap the target event, yet still select the wrong one. We study this scorer bottleneck on QVHighlights with a fixed QD-DETR-derived temporal expert: the parent-aware temporal expert reaches 64.71 R1@0.5, while diagnostic oracles over the same candidate family reach 75.68 R1@0.5 within the parent-reranked pool and 94.97 R1@0.5 over the full 100-candidate evidence pool. We introduce \ovetc{}, a rank-preserving counterfactual event verification layer that keeps the temporal expert's ranked list as the default and permits only high-confidence rank-1 overrides from a policy calibrated on held-out training videos. The verifier uses parent-aware interval evidence, structured Qwen3-VL event judgments, VideoLLaMA3 pairwise temporal signals, and dense temporal evidence, but never lets those signals freely reorder the top-$K$ list. On QVHighlights validation, \ovetc{} improves the parent-aware temporal expert from 64.71 to 65.35 R1@0.5 with a positive paired-bootstrap interval, while mAP is largely unchanged, from 42.67 to 42.74; it makes 20 overrides, with 15 helpful corrections and 3 harmful corrections. The results point to a practical constraint on verifier design: semantic compatibility is useful evidence, but temporal grounding still requires deciding where the event starts, changes, and ends inside a visually compatible scene.

\end{abstract}

\section{Introduction}
\label{sec:intro}
Video moment retrieval localizes the interval in a video that satisfies a natural-language query. In search, the useful answer is usually not the whole video, but the few seconds where the described event happens. Earlier temporal sentence grounding work studied this problem through proposal scoring, dense temporal maps, span prediction, local-global text-video interaction, and dense boundary regression \cite{gao2017tall,zhang2020twodtan,zhang2020vslnet,mun2020lgi,zeng2020drn,zhang2023survey}. QVHighlights made the problem especially visible by combining untrimmed web videos, free-form queries, relevant temporal moments, and saliency annotations \cite{lei2021qvhighlights}. Recent DETR-style systems improve temporal grounding through query-conditioned representations, foreground-background discrimination, length-aware matching, moment quantization, and more stable temporal query behavior \cite{carion2020detr,moon2023qddetr,moon2023cgdetr,park2024momentmix,sun2025mqvtg,tang2025simdetr}.

These advances are the reason we do not frame \ovetc{} as a leaderboard system. On the official QVHighlights test split, SG-DETR with pretraining reports 74.2 R@1@0.5, 60.4 R@1@0.7, and 58.8 average mAP \cite{gordeev2024sgdetr}; SMORE reports 76.39, 62.84, and 54.72 with an MLLM pipeline \cite{jeon2026smore}; and CVA reports 70.05, 55.32, and 47.49 with context-aware video-text alignment \cite{moon2026cva}. \ovetc{} is not a new temporal backbone. It asks a different question: when a temporal expert already produces a good candidate set, can a verifier correct the final selection without replacing the temporal model or collapsing the ranked list?

That question is different from training a larger first-stage model. Re-training a temporal architecture, adding video-language pretraining, or moving to an MLLM pipeline can improve absolute accuracy \cite{gordeev2024sgdetr,jeon2026smore,lu2024llavamr,xu2025momentgpt}, but it does not explain what happens after several plausible intervals have already been proposed. A post-hoc verifier must decide whether a candidate is better than a strong default, and it must do so without damaging the rest of the ranked list. We study that decision directly.

Our experiments show that many remaining errors are selection errors. A parent-aware QD-DETR temporal expert reaches 64.71 R1@0.5 and 42.67 mAP on QVHighlights validation, yet a diagnostic oracle over the same parent-reranked pool reaches 75.68 R1@0.5. An oracle over the full 100-candidate evidence pool reaches 94.97 R1@0.5. These oracles are not deployable methods. They show that the candidate pool often contains a much better answer than the selector chooses. The missing piece is not simply more candidates; it is calibrated selection among visually plausible alternatives.

The practical problem is that visually plausible intervals are often temporally wrong. An interval may contain the right person, object, scene, or coarse action while being too broad, shifted to the wrong repetition, or missing the event boundary. We tested this directly with Qwen3-VL counterfactual event verification. The verifier produced structured judgments about subject presence, action visibility, relations, before/during/after states, outside context, boundary transitions, and confidence. Used as an absolute ranker, however, it damaged performance: 45.42 R1@0.5 versus 64.71 for the temporal default. The VLM recognized semantic compatibility, but that compatibility was not calibrated for temporal selection.

\ovetc{} is built around this failure mode. It keeps the parent-aware temporal expert's ranked list as the default decision and uses additional evidence only as a conservative rank-1 override signal. The selector may replace the top prediction with an alternative when a policy calibrated on held-out training videos predicts a high-precision correction; ranks 2 and below preserve the parent-aware order. This rank-preserving design avoids the top-$K$ collapse seen in flat dense selectors, where many near-duplicate alternatives from one parent crowd out diverse candidates.

The rank-preserving constraint matches the strength of the evidence. The VLM and dense features can identify occasional mistakes in the top prediction, but they are not reliable enough to impose a new ordering over all candidates. We therefore treat verification as a counterfactual decision: would replacing the default top interval with this alternative likely improve temporal overlap? This keeps the claim narrow and testable.

\ovetc{} is deliberately sparse. On validation, it applies 20 overrides; 15 are helpful and 3 are harmful, giving 75.0\% helpful precision and an average override delta of +0.3367 tIoU. This improves the parent-aware temporal expert from 64.71 to 65.35 R1@0.5, with a paired-bootstrap 95\% interval of +0.258 to +1.097 percentage points. The mAP change is much smaller, 42.67 to 42.74, and its bootstrap interval crosses zero. R@5@0.5 remains 0.8516, with the same five distinct parents in the top five. These numbers support the main claim: verifier evidence helps when it is treated as a high-precision correction to a strong temporal default, not as an unconstrained ranking score.

The contributions are as follows. First, we provide an oracle-gap diagnosis showing that QVHighlights candidate pools can contain much better intervals than deployed selectors choose. Second, we introduce a reproducible counterfactual event verification pipeline with parent-aware interval evidence, precomputed VLM judgments, pairwise temporal signals, and dense temporal evidence. Third, we report negative results that clarify the failure mode: absolute Qwen3-VL scoring, conservative VideoLLaMA3 pairwise policies, and dense learned selectors do not reliably solve temporal selection on their own. Fourth, we introduce \ovetc{}, a rank-preserving consensus selector that improves top-1 recall by applying only high-confidence rank-1 corrections while preserving top-$K$ diversity.

\section{Related work}
\label{sec:related}
\paragraph{Temporal grounding before QVHighlights.}
Temporal sentence grounding has appeared under several names, including natural language video localization and video moment retrieval \cite{zhang2023survey}. The benchmark lineage includes cooking descriptions in TACoS \cite{regneri2013grounding}, personal-video moment localization in DiDeMo \cite{hendricks2017localizing}, dense event descriptions in ActivityNet Captions \cite{krishna2017dense}, and query- or importance-driven summarization benchmarks such as TVSum and YouTube Highlights \cite{song2015tvsum,sun2014ranking}. Early systems such as TALL/CTRL treated grounding as cross-modal proposal scoring and boundary regression \cite{gao2017tall}. Later methods used denser temporal structures or end-to-end predictors: 2D-TAN scores start and end pairs on a 2D temporal map \cite{zhang2020twodtan}, VSLNet casts localization as span prediction \cite{zhang2020vslnet}, LGI models local and global video-text interactions \cite{mun2020lgi}, and DRN uses dense regression supervision for boundary prediction \cite{zeng2020drn}. These works motivate our focus on interval selection. The hard part is choosing the span that best matches the sentence, rather than only recognizing query-related content.

\paragraph{Benchmarks and DETR-style moment retrieval.}
QVHighlights evaluates moment retrieval and highlight detection with natural-language queries, relevant temporal moments, and saliency labels \cite{lei2021qvhighlights}. Its Moment-DETR baseline adapted DETR-style set prediction \cite{carion2020detr} to video moments. Subsequent work improved the temporal representation, the decoder, or the training signal. QD-DETR conditions video features on the query \cite{moon2023qddetr}. UniVTG studies a unified formulation across temporal grounding tasks \cite{lin2023univtg}. TR-DETR models the reciprocal relation between moment retrieval and highlight detection \cite{sun2024trdetr}. CG-DETR uses fine-grained clip-word correlation for cross-modal calibration \cite{moon2023cgdetr}. SG-DETR adds saliency-guided cross attention, a hybrid detector, and InterVid-MR pretraining \cite{gordeev2024sgdetr}. FlashVTG and TDP-DETR focus on feature handling and temporal dynamics for joint moment retrieval and highlight detection \cite{cao2025flashvtg,an2026tdpdetr}. \ovetc{} operates after this kind of temporal expert. It asks how much selection quality can be recovered from a fixed expert's candidate pool.

\paragraph{Temporal discrimination, boundaries, and query conflicts.}
Several recent papers argue that temporal grounding errors come from weak discrimination between relevant and irrelevant moments, unstable alignment between global query semantics and local boundaries, or redundant proposal queries. Length-Aware DETR and MomentMix target short-moment failures by creating short-moment augmentations and conditioning matching on duration \cite{park2024momentmix}. BAM-DETR replaces center-based prediction with boundary-aligned decoding and quality-aware ranking \cite{lee2024bamdetr}. Sim-DETR analyzes why DETR-style designs behave differently in temporal sentence grounding than in object detection: similar target moments can create conflicts between queries, while a single query must represent both global semantics and local boundaries \cite{tang2025simdetr}. Moment Quantization improves relevant/irrelevant moment separation through a moment-level codebook \cite{sun2025mqvtg}. RGTR diversifies moment queries using explicit regional guidance \cite{sun2025rgtr}. CVA targets context-induced false negatives through query-aware context diversification and a boundary-focused contrastive loss \cite{moon2026cva}. These methods modify training, representation learning, or the decoder itself. \ovetc{} instead treats the trained model as a parent proposal expert and concentrates on the final selection step.

This separation matters for how the contribution should be read. Work on boundary-aware decoding and moment quantization changes the distribution of candidates produced by the model. Our setting keeps the candidate generator fixed and asks whether the final decision can be made more reliable by comparing nearby alternatives. In that sense, \ovetc{} is closer to an analysis of the decision surface than to a new temporal architecture. It complements backbone improvements: a stronger temporal expert could supply better parents, while the verifier would still need to decide when an alternative deserves to replace the default.

\paragraph{Reranking, diversity, and selective decisions.}
\ovetc{} is also related to work outside temporal grounding that separates scoring from final selection. Learning-to-rank methods such as RankNet learn pairwise preferences for ordered retrieval results \cite{burges2005ranknet}. Diversification methods such as maximal marginal relevance and xQuAD trade relevance against novelty so that a ranked list does not collapse to near-duplicate results \cite{carbonell1998mmr,santos2010xquad}. Detection pipelines face a related duplicate-control problem through non-maximum suppression and Soft-NMS, which adjust candidate scores according to overlap rather than treating every high-scoring proposal independently \cite{bodla2017softnms}. Selective classification studies a complementary question: when should a predictor abstain unless it can meet a desired risk level \cite{geifman2017selective}? \ovetc{} adopts this selective design principle, but the decision is not abstention. It is an asymmetric override: keep the temporal default unless an alternative clears a frozen high-precision gate.

\paragraph{Features and verification.}
Moment retrieval pipelines often include a scoring or reranking step, but the target of the score can vary. It may measure text-video compatibility, saliency, expected IoU, or confidence inherited from a detector. Frozen features also shape this scoring problem: CLIP provides transferable image-text alignment \cite{radford2021clip}, SlowFast captures visual motion at multiple temporal rates \cite{feichtenhofer2019slowfast}, and PANNs provide pretrained audio descriptors \cite{kong2020panns}. These signals can mix two questions: whether a window contains content related to the query, and whether it is the best temporal realization of the event. Our parent-aware formulation makes the second question explicit. It keeps the parent event hypothesis as the diversity unit, then verifies candidate intervals inside that parent using local contrast and boundary evidence.

The parent unit affects evaluation because top-$K$ moment retrieval depends on the tail of the ranked list. A flat reranker can improve one score while filling the list with small variations of the same temporal hypothesis. This is costly when the benchmark reports recall at multiple ranks or when downstream users inspect several candidate clips. Parent-aware verification separates two decisions that are often entangled: which coarse event hypothesis should remain in the ranked list, and which interval variant best expresses that hypothesis. \ovetc{} leaves the first decision mostly to the temporal expert and uses verification for the second only when the evidence is strong.

\paragraph{MLLMs and zero-shot temporal grounding.}
Large multimodal language models are useful for event verification because they can produce explicit judgments about objects, actions, relations, and temporal context. Video-ChatGPT, Video-LLaVA, and TimeChat adapt instruction-following models to video understanding and time-sensitive reasoning \cite{maaz2024videochatgpt,lin2024videollava,ren2024timechat}. Moment-GPT uses off-the-shelf MLLMs for zero-shot moment retrieval, combining query processing, candidate generation, and scoring \cite{xu2025momentgpt}. LLaVA-MR and SMORE adapt MLLM pipelines more directly to video moment retrieval through temporal encoding, frame selection, compression, or query-guided captions \cite{lu2024llavamr,jeon2026smore}. VideoLLaMA3 and Qwen3-VL are recent video-capable foundation models with reported gains in instruction following and visual reasoning \cite{zhang2025videollama3,bai2025qwen3vl}. These models motivate verifier-style grounding, especially when the query requires relational or state-based reasoning. However, semantic recognition does not by itself provide calibrated temporal ranking. Our Qwen3-VL diagnostic shows that a VLM can return structured judgments for every candidate and still choose poor intervals when used as an absolute ranker.

\paragraph{Structured event evidence.}
Training-free temporal grounding can decompose a query into sub-events, states, transitions, and order constraints without learning a new backbone \cite{zheng2024tfvtg}. Earlier action and situation-recognition work makes a related point: events are configurations of subjects, objects, relations, and roles unfolding over time \cite{zhu2016keyvolume,yatskar2016situation}. \ovetc{} uses this perspective in a constrained way. It evaluates each interval against nearby context and against a temporal default. The verifier is not a free-standing semantic judge; it decides whether an alternative interval has better event evidence than the default prediction.

\section{Method}
\label{sec:method}
\subsection{Problem setup and data flow}
Given a video $V$ and a natural-language query $q$, QVHighlights moment retrieval returns a ranked list of temporal intervals. A \emph{candidate interval} is a temporal proposal $c=[s,e]$ for the query-video pair. A \emph{candidate pool} is the fixed set of 100 candidate intervals considered before final selection. We use \emph{parent} for a coarse event hypothesis from the temporal expert and \emph{child} for a local boundary variant of that parent. A \emph{candidate instance} is one query-candidate pair together with its interval, semantic, and temporal evidence fields. Finally, an \emph{oracle} is a diagnostic selector that chooses the candidate with highest tIoU using ground truth after scoring; it is never a deployable method.

\ovetc{} assumes a temporal expert has already produced a ranked candidate list
\begin{equation}
  C = [c_1, c_2, \ldots, c_K] = g(V,q),
\end{equation}
where $c_1$ is the default top prediction. The goal is not to replace $g$, but to decide whether another candidate $c_i$ is sufficiently better than $c_1$ to justify an override. The predictor is
\begin{equation}
  \hat{C} =
  \begin{cases}
    [c_j, c_1, c_2, \ldots] & \text{if } G(E(c_j,c_1,q)) = 1,\\
    C & \text{otherwise,}
  \end{cases}
\end{equation}
with duplicate removal after an override. Here $E$ denotes interval, semantic, and temporal evidence, and $G$ is a frozen consensus rule selected on held-out training videos. The constraint is rank preservation: ranks below the top prediction keep the parent-aware temporal expert's order.

\begin{figure*}[t]
\begin{center}
\includegraphics[width=0.94\linewidth]{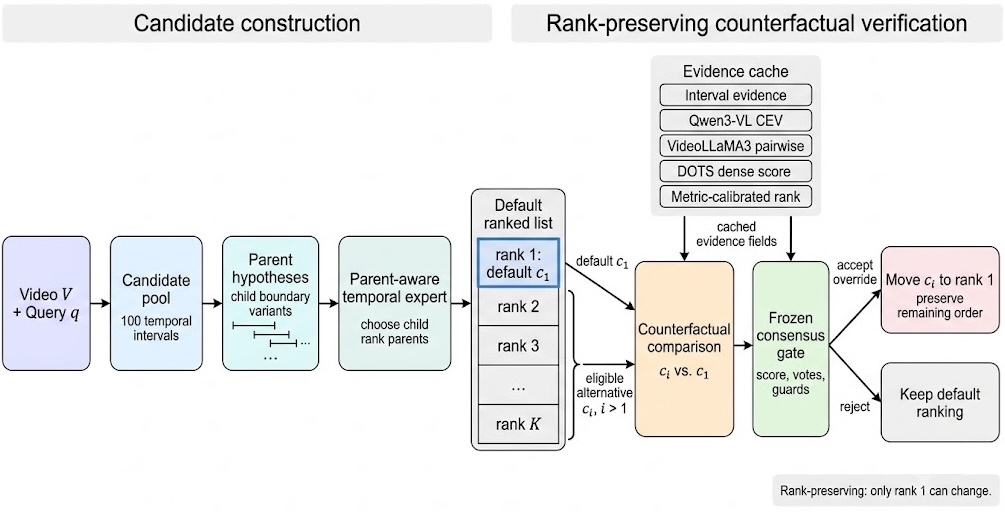}
\end{center}
\caption{\ovetc{} data flow. A fixed candidate pool is grouped into parent event hypotheses with child boundary variants. The parent-aware temporal expert selects one child per parent and supplies the default ranked list. Cached verifier evidence is used only for a counterfactual decision: whether an eligible alternative should replace the default top interval. If the consensus gate rejects the alternative, the original ranking is returned; if it accepts, the alternative is moved to rank 1 and the remaining parent-aware order is preserved.}
\label{fig:pipeline}
\end{figure*}

Figure~\ref{fig:pipeline} shows the data flow used by the rest of the method, and Table~\ref{tab:algorithm} gives the corresponding calibration and inference steps. Section~3.2 defines the candidate pool and parent-aware temporal expert. Section~3.3 attaches verifier evidence from Qwen3-VL, VideoLLaMA3, and dense temporal features. Section~3.4 describes the calibrated consensus gate that chooses whether to override rank 1. Section~3.5 states the leakage controls that keep diagnostic quantities separate from deployable features.

\begin{table*}
\begin{center}
\scriptsize
\setlength{\tabcolsep}{4pt}
\begin{tabular}{p{0.04\linewidth}p{0.91\linewidth}}
\hline
\multicolumn{2}{l}{\textbf{OVET-Consensus calibration and inference}} \\
\hline
Input & Video $V$, query $q$, temporal expert $g$, candidate-pool size $N=100$, evidence functions, frozen gate thresholds. \\
\hline
1 & Generate $N$ candidate intervals for $(V,q)$ and group them by parent event hypothesis. \\
2 & For each parent, score child boundary variants and keep the best child as the parent representative. \\
3 & Rank the parent representatives with the parent-aware temporal expert; let $c_1$ be the default top interval. \\
4 & For eligible alternatives, precompute or load Qwen3-VL CEV fields, VideoLLaMA3 pairwise fields, dense temporal scores, metric-calibrated ranks, and interval-contrast features. \\
5 & On held-out training videos only, select consensus thresholds under an override-rate constraint; freeze the policy before validation. \\
6 & At validation or test time, compare each eligible alternative with $c_1$ and compute its consensus score, vote count, and penalty count. \\
7 & If an alternative passes the frozen gate and guard constraints, move the highest-consensus alternative to rank 1 and remove duplicate intervals. \\
8 & Preserve the parent-aware order for all remaining ranks; if no alternative passes the gate, return the original ranked list. \\
\hline
\end{tabular}
\end{center}
\caption{Algorithm table for \ovetc{}. Calibration uses only training videos; QVHighlights validation labels are used only for reporting metrics.}
\label{tab:algorithm}
\end{table*}

\subsection{Parent-aware candidate foundation}
The candidate foundation is a QD-DETR-derived temporal expert \cite{moon2023qddetr}. Each parent represents a coarse event hypothesis, while its children are boundary variants obtained by local shifts, expansions, and interval refinements. Many children are near duplicates. If all children are ranked in a flat list, one plausible parent can occupy several top-$K$ positions and reduce candidate diversity.

The parent-aware temporal expert separates boundary choice from parent ordering. It first constructs the fixed pool of 100 candidate intervals per query. On QVHighlights this gives 721{,}800 training candidate instances and 155{,}000 validation candidate instances. Each candidate instance includes geometry, inherited parent rank and score, query-interval similarity, inside/outside contrast, boundary slopes, saliency mass, and CLIP, SlowFast, and PANN feature changes \cite{radford2021clip,feichtenhofer2019slowfast,kong2020panns}. Ground-truth windows, target tIoU, oracle labels, and hard-slice flags are excluded from deployable features.

The boundary selector chooses one interval per parent, and the parent reranker then orders the selected parent hypotheses. Both selectors are trained only on the training split with histogram-gradient-boosting regressors over interval and parent evidence. The parent-aware temporal expert is the default for \ovetc{} because it is stable: 0.5906 mean tIoU@1, 0.6471 R@1@0.5, 0.4774 R@1@0.7, and 0.8516 R@5@0.5 on validation.

The method uses these candidate instances in two different ways. For deployable scoring, they provide interval evidence and inherited temporal-model confidence. For diagnosis, the same candidate pool can be evaluated with an oracle after scoring. Keeping these roles separate lets us ask whether the candidate pool has enough coverage without leaking the answer into the selector.

\subsection{Multisource verifier evidence}
\ovetc{} augments the parent-aware temporal expert with precomputed counterfactual evidence. Here, counterfactual has a narrow meaning: the verifier asks whether an alternative candidate would be a better output than the default top prediction, not whether the alternative is plausible in isolation. We use this framing because many candidates contain the right entities or actions but still have the wrong temporal extent.

Qwen3-VL CEV \cite{bai2025qwen3vl} evaluates the top five candidate intervals per query with a strict JSON schema covering primary subject presence, auxiliary presence, action visibility, relation visibility, before-state match, during-event match, after-state match, outside-context match, boundary-transition visibility, and confidence. From these fields we derive event, contrast, transition, relation, and boundary scores. The evidence is semantic and structured, but our ablation shows that it is not a reliable absolute temporal ranker.

VideoLLaMA3 \cite{zhang2025videollama3} is used in a different form: a pairwise temporal comparator between the default prediction and selected alternatives, giving four comparisons per validation query. Its schema asks whether alternative $B$ is clearly more temporally precise than default $A$, with explicit fields for event match, temporal tightness, broader/shifted penalties, confidence, and an A/B/SAME choice. Pairwise-only policies selected no reliable override subset by themselves. We keep these outputs as evidence for the consensus gate rather than as a standalone method.

The third evidence source is our dense temporal selector, DOTS, trained over CLIP, SlowFast, PANN, and text/subtitle features. DOTS performs poorly when allowed to rerank the full candidate set, but its scores can still serve as weak evidence about candidate alternatives. A signal can be informative without being safe as a direct ranker.

All VLM evidence is precomputed before deployment-time scoring. This makes the experiments reproducible, avoids repeated model calls during threshold selection, and lets the same verifier outputs be evaluated under several decision rules. The verifier uses structured fields rather than free-form explanations, because a numerical gate needs stable evidence fields and explicit schema validation.

\subsection{Rank-preserving consensus}
\ovetc{} treats the parent-aware temporal expert as the default decision. For every eligible alternative to $c_1$, it builds pairwise consensus evidence: temporal-expert score deltas, QD score deltas, CEV event/contrast/transition/relation/boundary margins, VideoLLaMA3 pairwise fields, DOTS dense scores, metric-calibrated ranks, duration ratios, and source-rank differences. The gate is deterministic rather than a neural or sklearn classifier, and its thresholds are selected on a held-out split of the training videos.

The calibration split is video-level. We assign each video to one of five deterministic SHA1 buckets using the first eight hexadecimal digits, then hold out bucket 0 for policy search. This subset contains 1{,}462 training queries. Policy search maximizes the target metric family under an override-rate constraint, then freezes the rule before QVHighlights validation evaluation. Validation labels are therefore used only for reporting.

For candidate $c_i$, the policy computes a consensus score $s_i$, vote count $v_i$, and penalty count $p_i$ from a fixed linear rule over evidence margins, source rank, and duration ratio; missing margins are set to zero. The supplementary material gives the exact formula and vote/penalty triggers. In brief, votes require agreement among CEV, temporal-expert, DOTS, metric-calibrated, and VideoLLaMA3 signals. Penalties mark outside-context increases, overlong intervals, weak temporal support, low source rank, or pairwise broadness flags. The frozen policy accepts an override only if
\begin{equation}
  s_i \geq \tau_s,\quad v_i \geq \tau_v,\quad p_i \leq \tau_p,
\end{equation}
and if all guard constraints in Table~\ref{tab:consensus_policy} hold. The rule encodes the main design choice: semantic, dense, and pairwise signals can support a correction, but no single source can freely replace the temporal ranker.

We impose three constraints. First, validation labels are not used to set the final gate. Second, the gate may be sparse: the desired behavior is high precision, not many overrides. Third, ranks 2 and below preserve the parent-aware order. This protects the top-$K$ tail from the failure mode observed in flat candidate rerankers, where many similar variants from the same parent occupy multiple slots.

At inference time, we compare eligible alternatives with the top prediction using the evidence groups in Table~\ref{tab:evidence_groups}. In the frozen policy, accepted alternatives must have source rank at most 2 and duration ratio at most 1.15; Qwen3-VL evidence covers the top five candidates, and VideoLLaMA3 compares four default-vs-alternative pairs per query. If multiple alternatives pass, the highest-consensus alternative is moved to rank 1, duplicate intervals are removed, and all remaining ranks keep the parent-aware order. The method never re-sorts the tail by VLM or dense scores.

\begin{table}
\begin{center}
\scriptsize
\setlength{\tabcolsep}{4pt}
\begin{tabular}{lc}
\hline
Policy constraint & Selected value \\
\hline
Minimum consensus score $\tau_s$ & 8.747896 \\
Minimum votes $\tau_v$ & 6 \\
Maximum penalties $\tau_p$ & 0 \\
Maximum duration ratio & 1.15 \\
Maximum source rank & 2 \\
Minimum temporal-expert margin & -0.20 \\
Maximum outside-context increase & 0 \\
Maximum override rate & 0.03 \\
\hline
\end{tabular}
\end{center}
\caption{Frozen \ovetc{} policy selected on held-out training videos. The rule is deterministic at validation time; it is not fitted on QVHighlights validation labels.}
\label{tab:consensus_policy}
\end{table}

\begin{table}
\begin{center}
\scriptsize
\setlength{\tabcolsep}{3pt}
\begin{tabular}{lp{0.58\linewidth}}
\hline
Evidence group & Role in \ovetc{} \\
\hline
Parent-aware interval evidence & Default ranking, parent identity, boundary and contrast features. \\
Qwen3-VL CEV & Structured semantic/event plausibility and outside-context evidence. \\
VideoLLaMA3 pairwise & Conservative temporal comparison between default and alternative. \\
DOTS dense selector & Weak dense evidence from visual, audio, and text/subtitle features. \\
Metric-calibrated selector & Auxiliary metric-quality rank; \ovetc{} uses only whether this selector ranks the candidate first. \\
Consensus gate & Frozen training-calibrated rule to override rank 1 or preserve the default. \\
\hline
\end{tabular}
\end{center}
\caption{Evidence sources used by \ovetc{}. The method does not trust any source as an unconstrained ranker; each source contributes to a conservative override decision against the temporal default.}
\label{tab:evidence_groups}
\end{table}

\subsection{Leakage controls}
The experiments separate deployable scoring from diagnostics. Oracles are computed only after scoring to measure candidate-pool headroom. Deployable selectors do not use target tIoU, ground-truth windows, oracle labels, hard-slice flags, or any diagnostic feature derived from validation labels. Ground truth is used for evaluation and for training-calibration targets only. Diagnostic hard-slice features are excluded from model inputs because they are derived from target labels and would leak evaluation information.

\section{Experiments}
\label{sec:experiments}
\subsection{Evaluation settings and protocol}
The experiments separate four questions. Does the QD-DETR-derived candidate pool contain better intervals than the deployed selector chooses? Is absolute VLM verification calibrated enough to rank temporal candidates? Do dense learned selectors preserve the top-$K$ tail when allowed to rerank freely? Finally, can a rank-preserving consensus gate use verifier evidence safely by applying only high-precision rank-1 overrides?

We evaluate moment retrieval on QVHighlights validation \cite{lei2021qvhighlights}; saliency and highlight detection are outside scope. We report benchmark-style recall at IoU 0.5 and 0.7, average mAP, mAP at 0.5 and 0.75, and duration-split mAP for short, middle, and long moments. Diagnostics use mean tIoU@1 and recall at rank 1 or 5. Oracles use ground truth only after scoring to measure candidate-pool headroom and are labeled diagnostic upper bounds. For the final comparison, we also report paired bootstrap confidence intervals over query ids with 5{,}000 resamples.

Because this is a validation study, published test-set numbers serve as context rather than matched baselines. The methods in Table~\ref{tab:published_context} use different features, pretraining, and sometimes hidden test-server evaluation. QVHighlights performance is already high, and several recent methods improve the backbone, pretraining, or MLLM pipeline. \ovetc{} instead asks whether a fixed candidate source can be made more reliable at the selection step.

\begin{table*}
\begin{center}
\scriptsize
\setlength{\tabcolsep}{4pt}
\begin{tabular}{llccc}
\hline
Method & Reported setting & R@1@0.5 & R@1@0.7 & mAP Avg. \\
\hline
Moment-DETR \cite{lei2021qvhighlights} & test & 52.89 & 33.02 & 30.73 \\
QD-DETR \cite{moon2023qddetr} & test & 62.40 & 44.98 & 39.86 \\
CG-DETR \cite{moon2023cgdetr} & test & 65.43 & 48.38 & 42.86 \\
Sim-DETR \cite{tang2025simdetr} & test & 67.64 & 50.91 & 46.93 \\
FlashVTG \cite{cao2025flashvtg} & test, InternVideo2 & 70.69 & 53.96 & 52.00 \\
CVA \cite{moon2026cva} & test & 70.05 & 55.32 & 47.49 \\
SG-DETR w/ PT \cite{gordeev2024sgdetr} & test, InterVid-MR pretraining & 74.20 & 60.40 & 58.80 \\
SMORE \cite{jeon2026smore} & test, MLLM pipeline & 76.39 & 62.84 & 54.72 \\
\hline
\end{tabular}
\end{center}
\caption{Published QVHighlights test-set context from primary papers. Methods differ in features, pretraining, model scale, and whether they use MLLMs. \ovetc{} is evaluated below as a controlled validation method over a fixed QD-DETR-derived candidate source, not as a test-server leaderboard entry.}
\label{tab:published_context}
\end{table*}

Table~\ref{tab:published_context} is context, not a matched baseline table. The validation gain over a strong temporal default is only part of the result. Candidate availability, semantic verification, dense evidence, and final selection behave differently, so we analyze them separately.

\subsection{Main moment-retrieval results}
Table~\ref{tab:official_main} reports official-style QVHighlights moment-retrieval metrics on validation. \ovetc{} gives the highest R1@0.5 and the highest mAP point estimate among the listed deployable variants, while the metric-calibrated selector keeps the highest R1@0.7. The gain over the parent-aware temporal expert is conservative by design: it is obtained without sacrificing R@5 tail quality and with only 20 rank-1 overrides.

\begin{table*}
\begin{center}
\scriptsize
\setlength{\tabcolsep}{2.5pt}
\begin{tabular}{lcccccccc}
\hline
Method & R1@0.5 & R1@0.7 & mAP & mAP@0.5 & mAP@0.75 & short & middle & long \\
\hline
Feature-pack reranker & 63.87 & 47.87 & 41.86 & 63.57 & 42.06 & 8.30 & 43.92 & 49.39 \\
Geometry-only reranker & 63.94 & 47.94 & 42.19 & 63.69 & 42.66 & 6.49 & 45.19 & 49.73 \\
Parent-aware boundary selector & 64.00 & 47.94 & 42.57 & 64.10 & 43.06 & 6.96 & 45.61 & 49.86 \\
Parent-aware temporal expert & 64.71 & 47.74 & 42.67 & 64.34 & 43.09 & 7.17 & 45.82 & 49.50 \\
Metric-calibrated selector & 64.45 & \textbf{48.26} & 42.67 & 64.24 & 43.16 & 7.03 & 45.74 & \textbf{49.84} \\
\ovetc{} & \textbf{65.35} & 48.13 & \textbf{42.74} & \textbf{64.51} & \textbf{43.18} & \textbf{7.27} & \textbf{46.02} & 49.48 \\
\hline
\end{tabular}
\end{center}
\caption{Official-style QVHighlights validation metrics for moment retrieval. The table reports moment-retrieval metrics; saliency/highlight detection is outside the scope of this study. \ovetc{} gives the highest R1@0.5 and mAP point estimate among the listed deployable variants.}
\label{tab:official_main}
\end{table*}

Table~\ref{tab:bootstrap_ci} reports paired uncertainty for the deployable comparison. The R1 gains are small but positive under query-level bootstrapping. The mAP interval crosses zero, so we do not claim a statistically reliable mAP improvement; the safer reading is that \ovetc{} improves top-1 recall while leaving average precision nearly unchanged.

\begin{table}
\begin{center}
\scriptsize
\setlength{\tabcolsep}{5pt}
\begin{tabular}{lcc}
\hline
Metric & \ovetc{} $-$ parent-aware temporal expert & 95\% bootstrap CI \\
\hline
R1@0.5 & +0.645 pp & [0.258, 1.097] \\
R1@0.7 & +0.387 pp & [0.065, 0.774] \\
mAP & +0.0736 pp & [-0.0324, 0.1854] \\
mean tIoU@1 & +0.004345 & [0.001163, 0.007950] \\
\hline
\end{tabular}
\end{center}
\caption{Paired bootstrap confidence intervals over QVHighlights validation query ids using 5{,}000 resamples. R1 improvements are positive under this analysis, while the mAP difference is too small to treat as a reliable gain.}
\label{tab:bootstrap_ci}
\end{table}

\subsection{Oracle gap: the candidate pool is not the bottleneck}
Table~\ref{tab:oracle_gap} gives the main diagnostic. The parent-aware temporal expert reaches 0.5906 mean tIoU@1 and 0.6471 R@1@0.5, while a within-parent oracle over the same parent-reranked pool reaches 0.7023 and 0.7568. Table~\ref{tab:source_rank_oracle} shows the full-pool headroom: over the 100-candidate evidence pool, the oracle reaches 0.8583 mean tIoU@1, 0.9497 R@1@0.5, and 0.8200 R@1@0.7. In 1223 of 1550 validation queries, the oracle improves by at least 0.05 tIoU over the parent-aware boundary selector. This is the clearest evidence for the scorer-bottleneck thesis.

Not all of this headroom is open to a rank-preserving gate. Consider an oracle that keeps the default prediction unless an alternative from the evidence pool is better and also satisfies the gate's structural guards (source rank at most 2, duration ratio at most 1.15, temporal-expert margin at least $-0.20$, and no increase in outside-context match), checked before any evidence vote is counted. Because it can always keep the default, its values are not directly comparable with the rows of Table~\ref{tab:source_rank_oracle}. It reaches 0.7326 mean tIoU@1, 0.8006 R@1@0.5 and 0.6510 R@1@0.7. If only alternatives that pass the full frozen gate may replace the default, which happens for 20 queries, it reaches 0.5960, 0.6542 and 0.4819, close to the 0.5950, 0.6535 and 0.4813 of \ovetc{}. Candidate availability therefore does not imply that a correction can be identified at inference.

\begin{table*}
\begin{center}
\scriptsize
\setlength{\tabcolsep}{4pt}
\begin{tabular}{lcccc}
\hline
Method & mean tIoU@1 & R@1@0.5 & R@1@0.7 & R@5@0.5 \\
\hline
Parent-aware boundary selector & 0.5867 & 0.6400 & 0.4794 & 0.8510 \\
Parent-aware temporal expert & 0.5906 & 0.6471 & 0.4774 & 0.8516 \\
Within-parent oracle, parent order & 0.6724 & 0.7200 & 0.5865 & 0.9213 \\
Within-parent oracle, parent reranker & 0.7023 & 0.7568 & 0.6116 & 0.9355 \\
\hline
\end{tabular}
\end{center}
\caption{Diagnostic oracle gap over the parent-aware candidate pools. Oracles use ground truth only after scoring and are not deployable methods.}
\label{tab:oracle_gap}
\end{table*}

\begin{table}
\begin{center}
\scriptsize
\setlength{\tabcolsep}{4pt}
\begin{tabular}{lccc}
\hline
Candidate subset & mean tIoU@1 & R@1@0.5 & R@1@0.7 \\
\hline
source rank $\leq 1$ & 0.6613 & 0.7077 & 0.5768 \\
source rank $\leq 2$ & 0.7551 & 0.8206 & 0.6755 \\
source rank $\leq 3$ & 0.7875 & 0.8639 & 0.7181 \\
source rank $\leq 5$ & 0.8293 & 0.9142 & 0.7761 \\
source rank $\leq 10$ / full pool & 0.8583 & 0.9497 & 0.8200 \\
\hline
\end{tabular}
\end{center}
\caption{Full-pool diagnostic oracle by source-rank budget. The target interval is often close to the head of the candidate pool, but the learned selector still fails to choose it.}
\label{tab:source_rank_oracle}
\end{table}

\subsection{Hard-slice diagnosis}
The oracle gap is largest on cases where the top prediction fails but a good candidate exists near the head of the list. Table~\ref{tab:hard_slices} reports three diagnostic slices. In the most direct slice, \emph{hard\_top1\_fail\_top10\_success}, the parent-aware temporal expert reaches only 12.60 R@1@0.5, while the oracle over the same parent-reranked pool reaches 40.82. This is not a proposal failure: by construction, a successful candidate is available within the top 10 source candidates. It is a selection failure.

\begin{table*}
\begin{center}
\scriptsize
\setlength{\tabcolsep}{3pt}
\begin{tabular}{llcccc}
\hline
Slice & Method & Queries & mean tIoU@1 & R@1@0.5 & R@1@0.7 \\
\hline
top-1 fail, top-10 success & Feature-pack reranker & 365 & 0.2360 & 0.0904 & 0.0712 \\
top-1 fail, top-10 success & Parent-aware temporal expert & 365 & 0.2549 & 0.1260 & 0.0740 \\
top-1 fail, top-10 success & Diagnostic oracle & 365 & 0.4207 & 0.4082 & 0.2082 \\
\hline
boundary near miss & Parent-aware temporal expert & 1108 & 0.5734 & 0.5957 & 0.4359 \\
boundary near miss & Diagnostic oracle & 1108 & 0.6860 & 0.7301 & 0.5623 \\
\hline
short interval & Parent-aware temporal expert & 1439 & 0.5924 & 0.6546 & 0.4830 \\
short interval & Diagnostic oracle & 1439 & 0.7058 & 0.7637 & 0.6185 \\
\hline
\end{tabular}
\end{center}
\caption{Hard-slice diagnostics. The strongest evidence is the top-1-fail/top-10-success slice: a good candidate is available near the head of the list, but the selector does not choose it. Oracles are diagnostic upper bounds.}
\label{tab:hard_slices}
\end{table*}

These slices explain why verifier design matters beyond the aggregate gain. The hard examples are not uniformly hopeless; many contain a better candidate. The difficulty is that the better candidate is surrounded by near-misses that share the same scene, actor, or action context. In that setting, semantic compatibility can mislead a verifier.

\subsection{Negative verifier results}
Table~\ref{tab:negative} shows why \ovetc{} is rank-preserving. Qwen3-VL CEV returns structured event judgments, but absolute CEV ranking drops to 0.4603 mean tIoU@1. Manual delta gating asks whether an alternative improves the temporal default, but its precision is still too low for a deployable selector. VideoLLaMA3 pairwise policies are conservative and select no reliable override subset. DOTS shows the opposite failure mode: it reranks aggressively and damages R@5@0.5. These failures motivate the guarded consensus rule.

The Qwen3-VL result is useful because the model produces valid structured outputs. Its judgments are often semantically sensible. The problem is calibration. An interval can score well because the queried object and action are visible throughout a wider scene, while the ground-truth moment is a narrower transition inside that scene. This explains why the CEV oracle over the same evaluated candidate pool is strong but the absolute CEV ranker is weak: the pool has good candidates, but semantic scores do not order them correctly.

\begin{table*}
\begin{center}
\scriptsize
\setlength{\tabcolsep}{3pt}
\begin{tabular}{llcccc}
\hline
Variant & Decision rule & mean tIoU@1 & R@1@0.5 & R@1@0.7 & R@5@0.5 \\
\hline
Parent-aware temporal expert & temporal default & 0.5906 & 0.6471 & 0.4774 & 0.8516 \\
Qwen CEV ranker & absolute VLM score & 0.4603 & 0.4542 & 0.2690 & 0.8510 \\
Qwen confidence gate & absolute VLM + confidence & 0.4898 & 0.4942 & 0.3097 & 0.8510 \\
Manual CEV delta gate & pairwise scalar gate & 0.5911 & 0.6445 & 0.4742 & 0.8510 \\
VideoLLaMA3 pairwise & conservative comparator & 0.5867 & 0.6400 & 0.4794 & 0.8510 \\
Direct DOTS selector & dense full rerank & 0.5740 & 0.6142 & 0.4645 & 0.7258 \\
\ovetc{} & rank-preserving consensus & \textbf{0.5950} & \textbf{0.6535} & \textbf{0.4813} & \textbf{0.8516} \\
\hline
\end{tabular}
\end{center}
\caption{Ablation and negative findings. Absolute semantic rankers and dense full rerankers are unreliable as temporal selectors; \ovetc{} uses their evidence only through a rank-preserving consensus gate.}
\label{tab:negative}
\end{table*}

Table~\ref{tab:negative} is not a factorial leave-one-source-out ablation of \ovetc{}. These rows test each evidence family as an independent selector; they do not measure marginal contribution inside the final consensus rule. The result is still useful: no individual source is safe enough to replace the temporal expert. Section~\ref{sec:source_attribution} measures what each source contributes inside the rule.

\subsection{Source attribution inside the consensus rule}
\label{sec:source_attribution}
Table~\ref{tab:source_removal} removes one evidence source at a time from the frozen gate and recomputes the selection from cached evidence, with no VLM rerun, training, or tuning on validation. Qwen3-VL CEV is binding: without it no alternative clears the gate and the output equals the parent-aware temporal expert. DOTS supports most accepted corrections, since only 3 of the 20 overrides remain without it, and the metric-calibrated vote supports 4 of them. VideoLLaMA3 evidence is cached but changes none of the selected overrides at this operating point. Removing a source also changes which alternatives reach the vote threshold, so we read the table as an attribution audit rather than a causal decomposition. The table sharpens the negative result above: Qwen3-VL CEV is unsafe as a free ranker, yet the guarded rule depends on it.

\begin{table}
\begin{center}
\scriptsize
\setlength{\tabcolsep}{4pt}
\begin{tabular}{lcccccc}
\hline
Variant & R1@0.5 & R1@0.7 & mAP & Overrides & Helpful & Harmful \\
\hline
\ovetc{}, all sources & 65.35 & 48.13 & 42.74 & 20 & 15 & 3 \\
without Qwen3-VL CEV & 64.71 & 47.74 & 42.67 & 0 & 0 & 0 \\
without VideoLLaMA3 & 65.35 & 48.13 & 42.74 & 20 & 15 & 3 \\
without DOTS & 64.84 & 47.81 & 42.68 & 3 & 3 & 0 \\
without metric-calibrated vote & 65.29 & 48.13 & 42.74 & 16 & 12 & 2 \\
temporal and interval evidence only & 64.71 & 47.74 & 42.67 & 0 & 0 & 0 \\
\hline
\end{tabular}
\end{center}
\caption{Leave-one-source-out attribution inside the frozen consensus gate on QVHighlights validation, recomputed from cached evidence with no VLM rerun, training, or validation tuning. Rows with no override equal the parent-aware temporal expert. Helpful and harmful use $\pm 0.05$ tIoU relative to that expert.}
\label{tab:source_removal}
\end{table}

\subsection{Verifier scale and cost}
Table~\ref{tab:vlm_cost} reports the scale of the precomputed verifier evidence. The VLMs are not called at deployment-time scoring; their structured outputs are reused across policies. Qwen3-VL produces valid parsed JSON for all reported candidate evaluations. VideoLLaMA3 also has zero parse failures after pair-level deduplication, but its choices are overwhelmingly conservative, with only 9 validation pairs choosing the alternative candidate.

\begin{table*}
\begin{center}
\scriptsize
\setlength{\tabcolsep}{2.5pt}
\begin{tabular}{lp{0.12\linewidth}rrrrp{0.28\linewidth}}
\hline
Verifier & Unit/query & Train & Val & Total & Parse fail. & Evidence-generation cost \\
\hline
Qwen3-VL CEV & 5 candidates & 36{,}090 & 7{,}750 & 43{,}840 & 0\% & 4.8GB cache; val 4.9h, train 22.4h \\
VideoLLaMA3 pairwise & 4 pairs & 28{,}872 & 6{,}200 & 35{,}072 & 0\% & 30GB cache; pair-level deduplication \\
\hline
\end{tabular}
\end{center}
\caption{Precomputed VLM evidence used by the verifier. Evidence generation used one GPU; Qwen3-VL used 4 CPUs with 8GB per CPU, while VideoLLaMA3 used 8 CPUs with 10GB per CPU. Precomputation makes policy evaluation cheap, but the offline evidence generation cost is nontrivial.}
\label{tab:vlm_cost}
\end{table*}

\subsection{Override and diversity analysis}
\ovetc{} is sparse by design. Table~\ref{tab:override_analysis} shows that unconstrained or weakly gated verifier decisions produce many harmful overrides, while Table~\ref{tab:diversity} confirms that the accepted overrides preserve the top-$K$ tail. The consensus policy applies only 20 overrides, but 15 are helpful and 3 are harmful. Helpful and harmful use a $\pm 0.05$ tIoU band, whereas recall counts threshold crossings: the 20 overrides move 11 queries from fail to pass at R1@0.5 and 1 from pass to fail, and 7 and 1 at R1@0.7. This is the core design tradeoff: verifier evidence is useful only when it acts as a high-precision correction signal against a strong temporal default.

\begin{table*}
\begin{center}
\scriptsize
\setlength{\tabcolsep}{4pt}
\begin{tabular}{lcccccc}
\hline
Variant & Overrides & Helpful & Harmful & Neutral & Precision & Mean $\Delta$ tIoU \\
\hline
Raw CEV override policy & 927 & 241 & 608 & 78 & 0.2600 & -- \\
Manual CEV delta gate & 186 & 82 & 77 & 27 & 0.4409 & -- \\
VideoLLaMA3 guarded selector & 0 & 0 & 0 & 0 & -- & 0.0000 \\
\ovetc{} & 20 & 15 & 3 & 2 & 0.7500 & +0.3367 \\
\hline
\end{tabular}
\end{center}
\caption{Override analysis. Helpful and harmful are defined by $\pm 0.05$ tIoU relative to the parent-aware temporal expert. Direct verifier gates fire too often; \ovetc{} keeps only the high-consensus subset.}
\label{tab:override_analysis}
\end{table*}

\begin{table}
\begin{center}
\scriptsize
\setlength{\tabcolsep}{4pt}
\begin{tabular}{lccc}
\hline
Method & Unique parents@5 & Mean pairwise tIoU@5 & R@5@0.5 \\
\hline
Parent-aware temporal expert & 5.0000 & 0.1540 & 0.8516 \\
\ovetc{} & 5.0000 & 0.1540 & 0.8516 \\
\hline
\end{tabular}
\end{center}
\caption{Top-$K$ diversity analysis. \ovetc{} preserves the parent-aware tail exactly on validation.}
\label{tab:diversity}
\end{table}

Figure~\ref{fig:qualitative_examples} shows representative validation cases behind the override counts. The verifier can rescue a misplaced default, preserve a correct default, make a harmful high-confidence correction, and abstain when a diagnostic oracle candidate would have been better.

\IfFileExists{figures/qualitative_examples.png}{
\begin{figure*}[t]
\begin{center}
\includegraphics[width=0.68\linewidth]{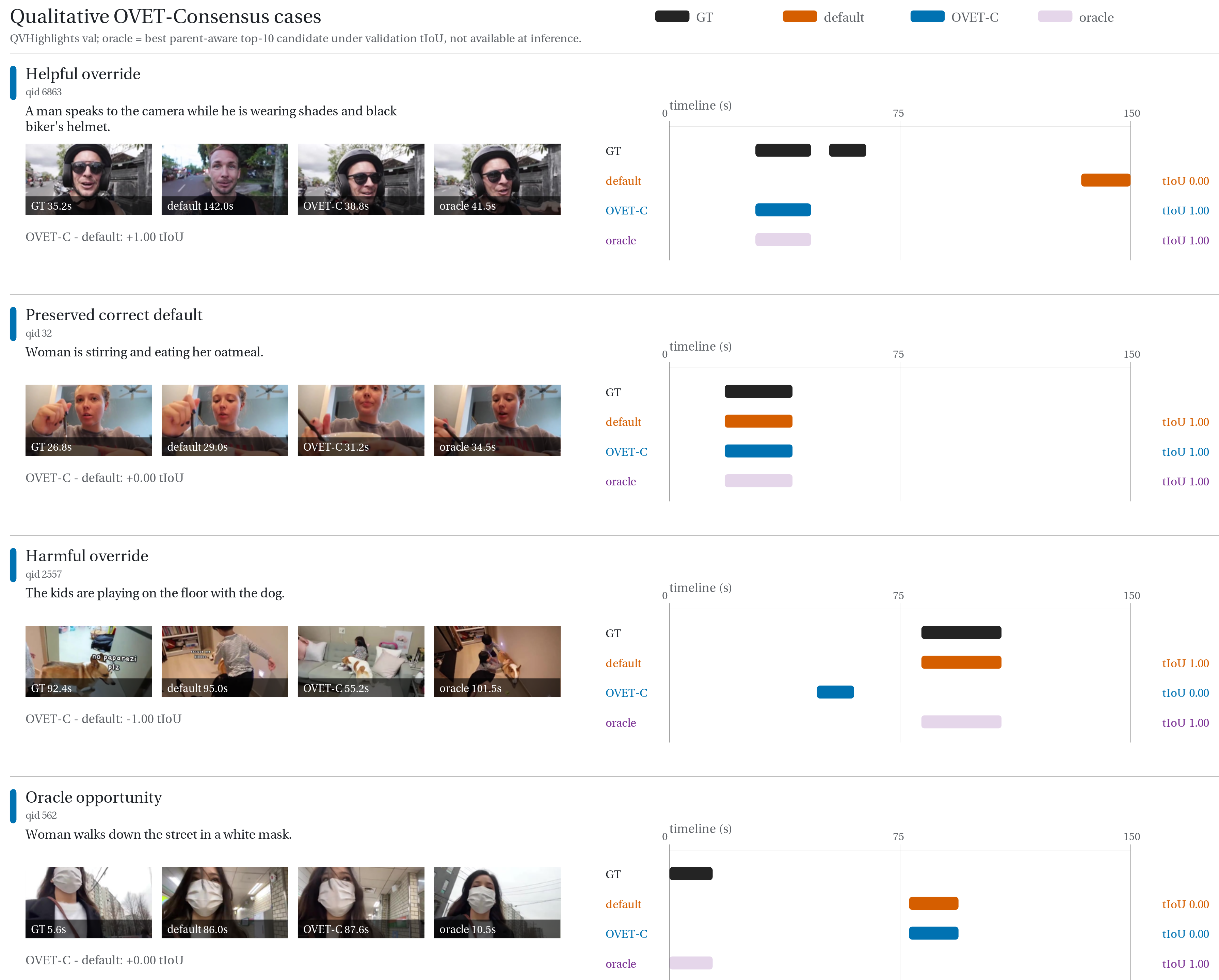}
\end{center}
\caption{Qualitative examples on QVHighlights validation. Times are relative to the 150-second clip. The diagnostic oracle is the best parent-aware top-10 candidate under validation tIoU and is not available to \ovetc{} at inference. Rows show a helpful override, a preserved correct default, a harmful override, and an abstained case where a better candidate exists.}
\label{fig:qualitative_examples}
\end{figure*}
}{}

The 20 accepted overrides are concentrated in shorter and middle-duration moments: 8 of 9 short-interval overrides are helpful, and 7 of 9 middle-interval overrides are helpful. The two long-interval overrides are not reliable. A narrow reading is appropriate: \ovetc{} acts as a sparse boundary-correction layer, improves R@1 without changing R@5, and leaves substantial oracle headroom.

\subsection{Gate sensitivity and calibration stability}
\label{sec:gate_sensitivity}
The gate is an operating point chosen for precision on held-out training videos, not for coverage. Table~\ref{tab:gate_sensitivity} changes one guard at a time and reports each variant on validation as a sensitivity check; the reported gate is not changed. Lowering the vote threshold from 6 to 5 more than doubles the overrides, from 20 to 46, and raises R1@0.5 to 65.55, but harmful overrides rise from 3 to 12 and precision falls from 0.750 to 0.543. Raising it to 7 keeps 11 overrides at precision 0.636. Changing the source-rank, duration or consensus-score limit moves R1@0.5 by at most 0.12.

We also re-select the gate on each of the five deterministic held-out training-video buckets; bucket 0 gave the reported gate (Appendix~\ref{app:repro}). The re-selected gates apply between 0 and 46 overrides. On validation, R1@0.5 stays between 64.71 and 65.42 and R1@0.7 between 47.74 and 48.13, never below the parent-aware temporal expert. Two buckets select a gate that never fires, and two select a looser gate with 46 overrides at precision 0.587. The gain therefore depends on the calibration subset: it is zero in those two buckets and never negative.

\begin{table}
\begin{center}
\scriptsize
\setlength{\tabcolsep}{3pt}
\begin{tabular}{lccccccc}
\hline
Gate & R1@0.5 & R1@0.7 & mAP & Overrides & Helpful & Harmful & Precision \\
\hline
reported gate & 65.35 & 48.13 & 42.74 & 20 & 15 & 3 & 0.750 \\
votes $\geq 5$ & 65.55 & 48.26 & 42.82 & 46 & 25 & 12 & 0.543 \\
votes $\geq 7$ & 65.03 & 48.00 & 42.70 & 11 & 7 & 3 & 0.636 \\
source rank $\leq 3$ & 65.42 & 48.13 & 42.74 & 23 & 16 & 4 & 0.696 \\
duration ratio $\leq 1.10$ & 65.23 & 48.06 & 42.72 & 18 & 13 & 3 & 0.722 \\
duration ratio $\leq 1.20$ & 65.42 & 48.13 & 42.74 & 21 & 16 & 3 & 0.762 \\
consensus score $\geq 8.248$ & 65.35 & 48.13 & 42.74 & 21 & 15 & 3 & 0.714 \\
consensus score $\geq 9.248$ & 65.35 & 48.13 & 42.74 & 19 & 15 & 2 & 0.789 \\
\hline
\end{tabular}
\end{center}
\caption{Gate sensitivity on QVHighlights validation. Each row changes one guard of the reported gate (votes $\geq 6$, source rank $\leq 2$, duration ratio $\leq 1.15$, consensus score $\geq 8.748$). Validation labels are used only to report these rows; the reported gate is unchanged. Precision is the share of overrides that are helpful.}
\label{tab:gate_sensitivity}
\end{table}
\FloatBarrier

\section{Discussion}
\label{sec:discussion}
We do not make a leaderboard claim. The deployable gain is small and concentrated in top-1 recall; the mAP change is not statistically reliable under our bootstrap analysis. The more useful result is the error decomposition: a large oracle gap remains after plausible candidates exist, and direct semantic or dense reranking does not close it. This fits the structure of moment retrieval, where the queried subject or action can appear throughout a longer compatible context. Rank preservation is a conservative response: \ovetc{} verifies only the top prediction, and if no alternative clears the gate, the temporal expert's list is returned unchanged.

\section{Limitations}
\label{sec:limitations}
\ovetc{} depends on the parent-aware temporal expert. If the candidate pool lacks a useful alternative, verification cannot recover the event, and the method is not a substitute for stronger temporal modeling, pretraining, or proposal generation. We report QVHighlights validation moment-retrieval metrics only, not hidden-test, cross-dataset, saliency, or highlight-detection results. VLM evidence is precomputed and costly, and Qwen3-VL and VideoLLaMA3 are verifier sources rather than standalone temporal rankers. The leave-one-source-out analysis (Table~\ref{tab:source_removal}) attributes accepted corrections to evidence sources but is not a causal decomposition. We leave temporal MMR, Soft-NMS suppression, and pairwise or listwise rankers to future work.

\section{Conclusion}
\label{sec:conclusion}
\ovetc{} treats verification as a guarded correction to a strong temporal default. On QVHighlights validation, the candidate pool often contains better intervals than the deployed selector chooses, while unconstrained semantic or dense reranking can damage the ranked list. A sparse, rank-preserving gate gives a small top-1 gain and leaves mAP nearly unchanged. The remaining oracle gap leaves a clear target: temporal verification that is better calibrated to event boundaries.

The evaluation lesson is simple: a final R@1 number does not say whether the system failed to propose the event or failed to choose it. Oracle gaps expose that distinction.

\clearpage
\bibliography{main}

\clearpage
\appendix
\section{Frozen Consensus Gate}
\label{app:gate}

The final \ovetc{} selector is deterministic rather than a learned classifier. For an alternative candidate $c_i$ and the default top prediction $c_1$, every margin is computed as alternative minus default. Missing margins are set to $0$. The consensus score is
\begin{align}
s_i ={}& 0.90v_i - 0.85p_i
 + 0.55m_{\mathrm{cev}}
 + 0.35m_{\mathrm{event}}
 + 0.35m_{\mathrm{contrast}} \nonumber\\
& + 0.30m_{\mathrm{temp}}
 + 0.28m_{\mathrm{dots}}
 + 0.22r_{\mathrm{pair}} \nonumber\\
& - 0.25\max(0, d_i-1.25)
 - 0.08\max(0, r_i-3),
\end{align}
where $v_i$ is the vote count, $p_i$ is the penalty count, $d_i$ is the candidate duration divided by the default duration, and $r_i$ is the candidate source rank. The terms $m_{\mathrm{cev}}$, $m_{\mathrm{event}}$, $m_{\mathrm{contrast}}$, $m_{\mathrm{temp}}$, and $m_{\mathrm{dots}}$ are CEV, event, contrast, temporal-expert, and DOTS margins. The term $r_{\mathrm{pair}}$ is the VideoLLaMA3 pairwise score.

\begin{table}[h]
\centering
\small
\begin{tabular}{p{0.78\linewidth}c}
\toprule
Vote condition & Votes \\
\midrule
CEV margin $\geq 0.75$ & 1 \\
Event margin $\geq 0.5$ or contrast margin $\geq 0.5$ & 1 \\
Temporal-expert margin $\geq -0.03$ & 1 \\
Source rank $\leq 3$ & 1 \\
Duration ratio in $[0.50,1.35]$ & 1 \\
DOTS score exists and DOTS margin $\geq 0.50$ & 1 \\
DOTS predicted tIoU is at least $0.03$ above the default DOTS predicted tIoU & 1 \\
Metric-calibrated selector rank is 1 & 1 \\
DOTS rank is 1 & 1 \\
VideoLLaMA3 strict-pass flag is active & 2 \\
VideoLLaMA3 chooses the alternative and pairwise score $\geq 1$ & 1 \\
\bottomrule
\end{tabular}
\caption{Vote definitions used by the frozen \ovetc{} gate.}
\end{table}

\begin{table}[h]
\centering
\small
\begin{tabular}{p{0.82\linewidth}c}
\toprule
Penalty condition & Penalties \\
\midrule
Outside-context match increases relative to the default & 1 \\
Duration ratio $>1.50$ & 1 \\
Source rank $>5$ & 1 \\
Temporal-expert margin $<-0.25$ & 1 \\
VideoLLaMA3 broader/shifted flag $\geq 2$ & 1 \\
\bottomrule
\end{tabular}
\caption{Penalty definitions used by the frozen \ovetc{} gate.}
\end{table}

An override candidate must satisfy all frozen guards:
\[
s_i \geq 8.747896,\quad
v_i \geq 6,\quad
p_i \leq 0,\quad
d_i \leq 1.15,\quad
r_i \leq 2,\quad
m_{\mathrm{temp}} \geq -0.20,
\]
and the outside-context increase must be at most $0$. The maximum override rate is $0.03$. If several candidates pass for the same query, the candidate with the largest consensus score is selected. Accepted query-level overrides are then sorted by consensus score and truncated to $\lfloor 0.03|\mathcal{Q}|\rfloor$. The selected candidate is moved to rank 1, duplicate interval signatures are removed, and the rest of the parent-aware ranking is preserved.

\section{Alternative Pool and Evidence Availability}

The default list comes from the parent-aware temporal expert. Candidate alternatives are merged from cached Qwen3-VL CEV rows, DOTS predictions, metric-calibrated predictions, and VideoLLaMA3 pairwise signals. Candidates with the same interval signature as the default are skipped.

Qwen3-VL CEV evaluates the top five candidate intervals per query. VideoLLaMA3 compares four default-vs-alternative pairs per query. DOTS and metric-calibrated predictions provide dense and metric-quality rank evidence, but neither source can reorder the output list directly in \ovetc{}.

\section{Metric-Calibrated Selector}

The metric-calibrated selector is an auxiliary quality-ranking source. In the final consensus gate it contributes only one vote: whether it ranks the alternative candidate first. Its quality score combines predicted overlap, helpfulness, harm risk, and temporal-model evidence:
\begin{align*}
&0.85\hat{t}
+0.42p_{0.5}
+0.50p_{0.7}
+0.18p_{\mathrm{helpful}}
-0.45p_{\mathrm{harmful}} \\
&+0.10\Delta
+0.08s_{\mathrm{interval}}
+0.04s_{\mathrm{QD}}
+0.03r^{-1}
+0.02\pi_d ,
\end{align*}
where $\hat{t}$ is predicted tIoU, $p_{0.5}$ and $p_{0.7}$ are predicted probabilities of clearing the two IoU thresholds, $\Delta$ is predicted improvement over the default, $s_{\mathrm{interval}}$ is the temporal-expert interval score, $s_{\mathrm{QD}}$ is the QD-DETR score, $r^{-1}$ is inverse source rank, and $\pi_d=-|\mathrm{duration\ ratio}-0.20|$.

\section{Caching and Reproducibility}
\label{app:repro}

Both verifier sources were generated offline and are reused by the frozen gate (Table~\ref{tab:vlm_cost}). The consensus policy is selected on a held-out training subset defined by deterministic video-level SHA1 buckets. Bucket 0 contains 1,462 training queries and is used for policy search. QVHighlights validation labels are used only for reporting.

\end{document}